\documentclass[journal,dvipsnames]{IEEEtran}
\usepackage{cite}

\ifCLASSINFOpdf
\else
\fi
\usepackage{amsmath}
\usepackage{url}

\usepackage{pgfplots} 
\usepackage[caption=false]{subfig}
\usepackage{multirow}
\usepackage{pict2e} 
\usepackage{sansmath}
\usepackage{booktabs}
\usepackage{tablefootnote}
\usepackage{multirow}
\usepackage{booktabs}
\usepackage{pifont}

\usepackage{float}
\restylefloat{table}

\usepackage[hidelinks]{hyperref}

\usepackage{calc} 
\usepackage{xurl}

\usepackage{color}

\newcommand{\newtext}[1]{\textcolor{NavyBlue}{#1}}
\renewcommand{\newtext}[1]{\textcolor[rgb]{0,0.0,0.0}{#1}}

\newcommand{\figref}[1]{Fig.~\ref{#1}}    
\newcommand{\Figref}[1]{Figure~\ref{#1}}  
\newcommand{\tabref}[1]{Table~\ref{#1}}

\newcommand{\secref}[1]{\S\ref{#1}}

\newcommand{\appref}[1]{Appendix~\ref{#1}}

\newcommand{\eg}{e.g., }

\newcommand{\ie}{i.e., }
\newcommand{\vs}{vs.\ }

\newcommand{\delete}[1]{}
\newcommand{\cmark}{\ding{51}}%
\newcommand\dsname{EduQG}
\newcommand\mrsc{MRS-count}
\newcommand\mrc{MRS}

\usepackage{amsmath} 
\usepackage[inline,shortlabels]{enumitem}

\usepackage[colorinlistoftodos,prependcaption,textsize=normalsize]{todonotes}

\usepackage{graphicx}
\graphicspath{ {./figures/} }

\usepackage[mathscr]{euscript}

\DeclareSymbolFont{rsfs}{U}{rsfs}{m}{n}
\DeclareSymbolFontAlphabet{\mathscrsfs}{rsfs}

\usepackage{amssymb}

\usepackage{multirow}
\usepackage{enumitem}

\begin{document}
%
\title{
EduQG: A Multi-format Multiple Choice Dataset\\for the Educational Domain
}
%
%
%

\author{Amir~Hadifar,
        Semere~Kiros~Bitew,
        Johannes~Deleu,\\
        Chris~Develder~\IEEEmembership{Senior Member,~IEEE,}
        and~Thomas~Demeester
        \textbf{}

\thanks{The authors are with the Text-to-Knowledge (T2K) team at IDLab, Ghent University-imec, Belgium. (e-mail: {amir.hadifar, firstname.lastname}@ugent.be)}
}

%
%

\markboth{IEEE Transactions on Learning Technologies, Vol.~XX, No.~X, XXXXXX~20XX}%
{Shell \MakeLowercase{\textit{et al.}}: Bare Demo of IEEEtran.cls for IEEE Journals}
%



\maketitle

\begin{abstract}
We introduce a high-quality dataset that contains 3,397 samples comprising \begin{enumerate*}[(i)]
\item multiple choice questions,
\item answers (including distractors), and
\item their source  documents,
\end{enumerate*}
from the educational domain.
Each question is phrased in two forms, normal and cloze.
Correct answers are linked to source documents with sentence-level annotations.
Thus, our versatile dataset can be used for both question and distractor generation, as well as to explore new challenges such as question format conversion.
Furthermore, 903 questions are accompanied by their cognitive complexity level as per Bloom's taxonomy.
All questions have been generated by educational experts rather than crowd workers to ensure they are maintaining educational and learning standards.
Our analysis and experiments suggest distinguishable differences between our dataset and commonly used ones for question generation for educational purposes.
We believe this new dataset can serve as a valuable resource for research and evaluation in the educational domain.
The dataset and baselines will be released to support further research in question generation.\footnote{\url{https://github.com/hadifar/} - upon acceptance.}
\end{abstract}

\begin{IEEEkeywords}
Question generation, multiple choice question, natural language processing, online learning.
\end{IEEEkeywords}

%
\IEEEpeerreviewmaketitle

\section{Introduction}
\label{sec:intro}

\IEEEPARstart{F}{rom} the time of Socrates to the present day, questions have been used as an effective teaching technique to facilitate and evaluate comprehension.
However, devising high-quality questions has always been challenging and time-consuming due to the extensive human domain knowledge required and the extra effort needed to adapt it to individuals.
\cite{davis2009tools} pointed out that even professional test developers \delete{could}do not manage to write more than three or four good Multiple Choice Questions (MCQs) per day.
Moreover, \delete{the }correction is labor-intensive
\delete{when administered }in a large group setting and may result in delayed feedback to students, especially when multiple graders are involved \cite{kim2012incorporation}.
Consequently, researchers proposed automatic approaches\delete{to smooth the way for creation and correction process} to facilitate more efficient question construction and correction.

\begin{figure}[t!]
    \centering
    \includegraphics[width=\columnwidth]{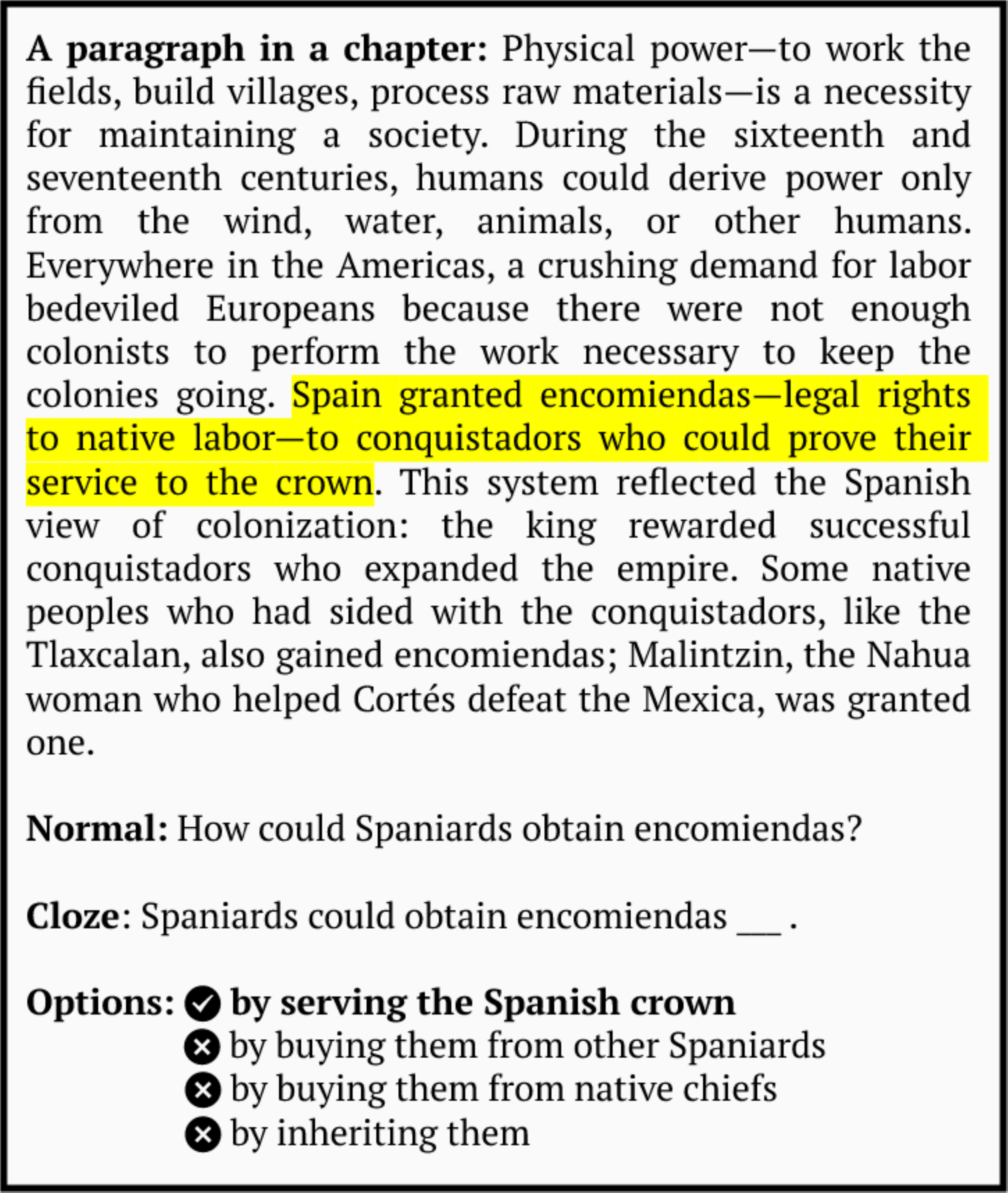}
    \caption{Example entry of \dsname{}. 
    For each entry two formats of the same question has provided (Normal and Cloze), in addition to the answer-key and sentence-level annotation.}
    \label{fig:example_of_q}
\end{figure}

For the \emph{construction} of questions, researchers developed Question Generation (QG) systems.
The input of QG systems typically comprise sentences, paragraphs, documents, tables, or images, from which a set of questions is generated.
These systems have been used in various ways for educational purposes, such as dialog systems (\eg \cite{lane2005teaching} suggested a dialog\delete{ue-based} system for novice programmers to help them plan and learn the tacit knowledge of programming) and reading tutor systems (\eg \cite{alsubait2016ontology} presented a QG system that was designed to help children in grades 1-3 to better understand a text).

For more efficient question \emph{correction}, MCQs were devised.
An MCQ contains a stem (\ie the question itself), the true answer, as well as distractors.
The correction process today is quite straightforward, \eg using optical mark recognition machines or similar computer vision solutions.
As a result, the challenging part that remains to be solved is MCQ generation.
The conventional approach to creating an MCQ is to employ a Distractor Generation (DG) model on top of a QG model.
To generate distractors, a DG system receives the question, true answer, and possibly source document text \cite{gao2019distractor}.

Although automatic MCQ generation systems have been around for a while, most publicly available datasets
do not suffice to build systems of sufficient quality for 
student assessment in the educational domain.
Existing MCQ datasets were mainly designed for Question Answering (QA) and created by crowdworkers rather than teachers or educational experts \cite{chen2018learningq}.
Teachers' questions typically serve formative and summative assessment needs \cite{marbach2000can} rather than merely evaluating students' recall skills.
Moreover, to achieve trust and adoption by educators, MCQs need to be properly grounded in the source documents --- most current MCQ datasets lack such fine- or coarse-grained annotation of source texts.

In this work, we
construct a new educational question generation (QG) dataset, {\dsname} (\secref{sec:dataset}), that contains 3,397 multiple choice questions (analyzed in detail in \secref{sec:data_analysis}).
The \dsname{} dataset can be used
to finetune existing QG models, as well as benchmarking QG for education (\secref{sec:baseline_models} provides baseline models and their performance).
As illustrated in \figref{fig:example_of_q}, in each MCQ sample the question is phrased both as a normal question and a cloze form (\ie fill-the-blank) thereof.
This \emph{multi-format} schema is not only valuable from the natural language processing perspective (cf.~the link between the cloze form and masked language models) but also for educational purposes \cite{holley1993relationship,reardon2018relationship,stanger2012multiple,martinez1999cognition,baldauf1979matching,gellert2013cloze}. 
For example, some studies showed that cloze format questions are preferable to other formats for assessing specific types of knowledge such as reading comprehension and grammar \cite{baldauf1979matching,gellert2013cloze}.   
\cite{reardon2018relationship} analyzed test scores for 8 million students and discovered that boys tend to outperform girls on MCQs while girls performed significantly better on open-ended questions. Also, it was shown that varying formats in assessments induce different approaches to problem-solving and learning \cite{holley1993relationship}.

Additionally, we annotated the key sentence(s) in the source text (highlighted in \figref{fig:example_of_q}), required to answer the question. Or, inversely, a QG system could in theory generate the question from that key sentence in the context of the source text.
Besides, 903 of the questions are accompanied by their cognitive complexity level according to Bloom's taxonomy \cite{anderson2001taxonomy}, which facilitates performance analysis of QG and DG models in function of these question types of varying complexity.

\section{Related Work}
\label{sec:related_work}
An early notion of QG goes back to the 70s when researchers altered a declarative sentence into an interrogative one by a set of syntactic transformations \cite{Wolfe1976AutomaticQG}. They improved this idea by using more sophisticated feature engineering or handcrafted templates. For example, \cite{heilman2009question} used a constituency parser to produce syntactic trees and a set of transformation rules to create questions. Similarly, \cite{kalady2010natural} included name-entity recognition, and keyword extraction to enhance this syntactic-style transformation. Some researchers suggested template-based methods where pre-defined templates are used to generate questions \cite{mostow2009generating,mazidi2015leveraging}. For instance, \cite{mazidi2015leveraging} constructed 50 question templates and utilized a semantic-role-labeling and dependency parser to find the corresponding template for a given input sentence to form a question. While some researchers partially created their own QG dataset \cite{smith2008question}, thanks to the syntactic nature of the methods, these datasets were mainly used for evaluation purposes rather than training question transformation objectives.

In recent years, there has been a significant shift to sequence-to-sequence models in which a model generates a set of questions given a text segment. As these models depend greatly on training data, having access to a suitable training set is critical.
However, due to the unavailability of large-scale training datasets designed for QG and because of the duality of the QG task with QA \cite{tang2017question}, researchers have adopted the available QA datasets for training QG models. \cite{su2020multi,wang2018qg}. In the next section (\secref{sec:non_edu}), we will overview some general purpose QA datasets and then in \secref{sec:edu} will look into existing datasets in the educational domain. 

\subsection{Non-Educational question datasets}
\label{sec:non_edu}
There is a sizeable body of works (see, \eg \cite{rogers2021qa,kurdi2020systematic} and the references therein) that focus on QA and QG. The majority of QA datasets are centered around one of these two aspects: \begin{enumerate*}[(i)]
    \item \textit{information-seeking} where the questioner did not know the answer e.g., the questioner submit a query in a search engine to find the answer  
    \item \textit{knowledge-probing} where the questioner intended to test knowledge of another person or machine e.g., the questioner is a teacher and answerer is a student.
\end{enumerate*}

Two famous examples in the \textit{information-seeking} category are  NaturalQ \cite{kwiatkowski2019natural} and  MS-Marco \cite{nguyen2016ms} where questions are generated by internet-users and paired with relevant document(s). Although these datasets have been used for QG, they are less suitable for the educational setting due to noise and format (users practically never posed a question in multiple-choice or cloze format). The second category, \textit{knowledge-probing}, is more popular among QG researchers with the famous examples of SQuAD \cite{rajpurkar2016squad} and HotpotQA \cite{yang2018hotpotqa}. While the two previously mentioned datasets rely on Wikipedia as a source of information, some relevant researches exist in other domains such NewsQA \cite{trischler2016newsqa} which is a QA dataset on news articles,  QUASAR-S \cite{dhingra2017quasar} which is a collection of cloze questions constructed from definitions of software entity tags, or SWAG \cite{zellers2018swag} which is a collection of MCQs, generated from video captions.

Both categories have been utilized for the task of QG, and results suggest that both appear promising for factoid QG. However, their limitations for the educational domain were recently pointed out 
\cite{chen2018learningq}. These datasets are mainly proposed for testing machines rather than humans, generated by crowdworkers rather than teachers, heavily focusing on name entities and mostly targeting recalling skills. Moreover, their domain was based on Wikipedia or news articles rather than educational textbooks,
which is quite different in terms of technical vocabulary, abstract nouns, complex sentences, and ordinary words used in non-typical ways \cite{espin2013curriculum}.

\subsection{Educational question datasets}
\label{sec:edu}
The above-mentioned limitations tried to address in different ways in the recent past. Researchers proposed datasets that originated from materials written by educators to test students. For example, RACE-C  \cite{pmlr-v101-liang19a} and CLOTH \cite{xie2017large} introduced multiple-choice reading comprehension datasets, collected from English examinations. ReClor \cite{yu2020reclor} is multi-choice reading comprehension questions and extracted the data from standardized graduate admission examinations (\eg GMAT). 
 SciQ \cite{welbl2017crowdsourcing}, TabMCQ \cite{jauhar2016tabmcq} and OpenBookQA \cite{mihaylov2018can} obtained MCQs from scientific contents, but the questions were written by crowdworkers. It is simply not possible to write a good MCQ in a short limited time \cite{davis2009tools,horbach2020linguistic} which is the case in crowdsourcing annotations (e.g., TabMCQ questions were created in approximately 70 seconds as reported by authors). Moreover, SciQ distractors are generated automatically followed by a post-filtering by crowdworkers that makes it less plausible for student's assessment.
 
 

ARC \cite{clark2018think} and TQA \cite{kembhavi2017you} are collections of MCQs for students. Both datasets only provide question-specific (i.e., answer-key) annotations, and the lack of explicit alignment with course content (e.g., on the sentence or paragraph level), hinders their 
usage for QG. Similarly, LearningQ \cite{chen2018learningq} introduced a collection of questions where obtained from online teaching platforms. Although this collection contains a large pool of question-document pairs, the absence of fine-grained and coarse-grained answer excerpts only allows for answer-agnostic QG explorations. Closely related to our work, although more narrow in scope (biology questions only) and size (585 questions in total), is ProcessBank \cite{berant2014modeling} which offers a collection of binary choice questions. 

The comparison between \dsname{} and some popular datasets is summarized in \tabref{tab:qg_datasets}. We divided the table in educational \vs non-educational and compared them regarding: 
\begin{enumerate*}[(i)]
\item source of creation (\emph{creator})
and question format (\emph{normal} or \emph{cloze}),
\item its answer type, categorized as 
Multiple-Choice (\emph{MC}) 
and/or \emph{extractive} (denoting that spans or entire sentences are considered
as answer), and
\item the \emph{type} and \emph{domain} of their \emph{source text} (e.g., paragraphs vs.~entire documents, originating from texbooks or Wikipedia articles). 
\end{enumerate*}


\begin{table*}
    \centering
    \renewcommand{\arraystretch}{1.2}
    \footnotesize
    \caption{\label{tab:qg_datasets} Qualitative comparison of datasets.}
    \begin{tabular}{c l ccc cc cc}
    \toprule

    & &
    \multicolumn{3}{c}{\textbf{question}} &
    \multicolumn{2}{c}{\textbf{answer}} &
    \multicolumn{2}{c}{\textbf{source text}}
    \\
    \cmidrule(r{3pt}){3-5}
    \cmidrule(l{3pt}r{3pt}){6-7}
    \cmidrule(l{3pt}){8-9}


    & {\textbf{Dataset}} &
    
    
    \multicolumn{1}{c}{\textbf{creator}} & \multicolumn{1}{c}{\textbf{normal}} &
    
    \multicolumn{1}{c}{\textbf{cloze}} & \multicolumn{1}{c}{\textbf{MC}} & 
    
    \multicolumn{1}{c}{\textbf{extractive}} &
    \multicolumn{1}{c}{\textbf{type}} & \multicolumn{1}{c}{\textbf{domain}}\\
    \midrule
    
    \multirow{6}{*}{\rotatebox{90}{Non-educational}} 
    & SQuAD    & crowd & \cmark & -      & - & \cmark & paragraph & Wikipedia  \\
    & HotpotQA & crowd & \cmark & -      & - & \cmark  & paragraph & Wikipedia  \\
    & NaturalQ & crowd & \cmark & -      & - & \cmark  & document & Wikipedia  \\
    & NewsQA   & crowd & \cmark & -      & - & \cmark & document & News  \\
    & SWAG & crowd & \cmark & - & \cmark & - & - & Video captions \\
    & QUASAR-S & auto & -      & \cmark & - & \cmark  & paragraph & Stackoverflow  \\
    
    \midrule

    \multirow{10}{*}{\rotatebox{90}{Educational}}  
    
    
    & CLOTH       & educator & - & \cmark & \cmark & \cmark  & paragraph & Standardized tests   \\
    & RACE-C      & educator & -$^{[1]}$ & \cmark & \cmark & \cmark  & paragraph & Standardized tests  \\
    & ReClor      & educator & \cmark & - & \cmark & -  & paragraph &  Standardized tests  \\
    & LearningQ   & educator & \cmark & - & - & -   & document & Online courses   \\
    & ARC         & educator & \cmark & - &  \cmark & -  & - & textbook  \\
    & OpenBookQA  & crowd    &\cmark & - &  \cmark & \cmark  & paragraph & textbook  \\
    & TQA       & educator    & \cmark & - & \cmark & -   & document & textbook  \\
    
    & SciQ        & crowd    & \cmark & - & \cmark & {\scriptsize \cmark \par}$^{[2]}$    & paragraph & textbook  \\
    & ProcessBank & educator & \cmark & - & \cmark & \cmark   & paragraph & textbook   \\
    & \dsname{}\quad\emph{(Ours)} & educator & \cmark & \cmark & \cmark & \cmark  & document & textbook \\ 
    \bottomrule
    \multicolumn{9}{l}{\footnotesize ${[1]}$ If questions are either \emph{normal} or \emph{cloze} form, the majority is chosen as representative of a column.} \\
    \multicolumn{9}{l}{\footnotesize ${[2]}$ SciQ provides answers in the form of paragraphs rather than spans or sentences.}

    \end{tabular}
    \begin{flushleft}
    \end{flushleft}
    %
\end{table*}

\section{\dsname{} Development}
\label{sec:dataset}


To develop our {\dsname} dataset, we chose to work with educational texts and related questions from Openstax,\footnote{\url{https://openstax.org}} which offers free textbooks and questions that have been developed and peer-reviewed by educators. 
We crawled all English contents\footnote{Available under a Creative Commons License.} using Openstax's public API, which resulted in a set of 43,578 questions that we then further filtered.
Indeed, since we aimed for questions that do not require mathematical reasoning, we excluded\footnote{\appref{app:list_of_books} lists all Openstax books that we retained.} topics such as physics, statistics, and algebra.
We further kept only multiple choice questions (MCQs), which is a common strategy in educational datasets (see \tabref{tab:qg_datasets}): well-designed MCQs allow measuring diverse types of knowledge, skills, and competences \cite{shin2019multiple, davis2009tools}.
Our final {\dsname} dataset thus amounts to 5,018 questions, which are related to 13 books (283 chapters) and comprise 3,493 \emph{normal} plain text questions and 1,525 in \emph{cloze} form.

We further enriched this crawled question set with 
\begin{enumerate*}[(i)]
\item \label{it:grounding} grounding passages that support the correct answer (detailed in \secref{sec:answer_sentence_selection}),
\item \label{it:cloze} cloze forms of the plain text questions and vice versa (see \secref{sec:cloze_generation}), and
\item Bloom's taxonomy complexity levels (see \secref{sec:bloom_taxo}).
\end{enumerate*}
For \ref{it:grounding}--\ref{it:cloze}, we employed trained annotators.\footnote{We hired two master's students in linguistics with prior experience in annotation projects, compensated at 13.5 EUR/h.}

\subsection{Answer Selection}
\label{sec:answer_sentence_selection}
To support the development of either answer selection, or question generation models, we collected for each question annotations of the grounding passage(s) in the source texts.
To this end, we facilitated the annotators with the automated retrieval of paragraphs from the considered chapter, to allow them 
to quickly select the correct passage from a list of suggestions rather than having to go through the full text.
This was implemented in a web platform, with multiple stages:\footnote{Further details and screenshots of the annotation interface are provided in \appref{app:annotation_platform}.}
\begin{enumerate*}[(i)]
\item \label{it:annot:list} we list the question, and the paragraphs of the corresponding chapter, which the annotator can choose to view  either in original order or as ranked by BM25 \cite{trotman2014improvements} (using as a query the concatenation of question and correct answer);
\item  \label{it:annot:paragraph} then the annotator selects a paragraph that fully/partially contains information leading to the correct answer;
\item  \label{it:annot:sentence} for a selected paragraph, the annotator subsequently selects one or more (and no more than strictly needed) sentences comprising the answer information;
\item we repeat steps \ref{it:annot:paragraph}--\ref{it:annot:sentence} to allow the annotator to indicate multiple supporting passages (for at most 15\,min per question).
\end{enumerate*}

We further applied a new post-processing step to check the questions that were flagged by the annotators as problematic, and filter them away if necessary. The latter 
questions mainly fall into four categories: \begin{enumerate*}[1)]
    \item arithmetic questions (\eg requiring mathematical reasoning)
    \item non-factoid questions (\eg requiring a complex answer such as opinion)
    \item media-related questions (\eg requiring a plot to answer the question).
    \item un-answerable questions (\eg annotators could not find the answer in the given time)
\end{enumerate*}.
The retained collection after post-processing contains 3,397 MCQs, of which 1,356 are in \textit{cloze} form and 2,041 in \textit{normal} form. 

To measure annotation agreement we randomly sampled 125 questions and assigned them to two annotators.

For the higher level annotation (\ie support/partial vs. no-support), annotators display an agreement of 90.4\% of the paragraphs, with a Kappa score of 0.8. For the finer level annotation (\ie sentence selection), annotators fully agreed on selecting the same set of sentences
for 42.9\% of the paragraphs, while agreement on at least one common sentence between two selections (\ie full or partial agreement) is 76.3\%. 
For 23.7\% of paragraphs we found no agreement between annotators. We hypothesize that this is due to the fact that frequently multiple distinct (sets of) sentences in a chapter allow answering the same question.
It should be noted that calculating Kappa scores for comparing the sentence selection appeared less straightforward, and therefore we decided to only report the agreement percentage instead.


\subsection{Question Generation}
\label{sec:cloze_generation}
In the second phase of the annotation process, we further enriched the dataset. 
For each question, we created a semantically identical but structurally different counterpart. More specifically, we hired two linguistic experts to write a \textit{cloze} formulation of a given MCQ from its \textit{normal} formulation. 
This conversion not only renders the dataset more homogeneous; it 
also gives us the opportunity to tackle 
the new task of question format conversion (see \secref{sec:format_conversion}), with potential practical value for e-assessment systems.
The experts converted a \textit{normal} MCQ to its corresponding \textit{cloze} formulation by considering two rules: \begin{enumerate*}[i)]
    \item The gap is replaceable by all candidate answers, which then leads to a grammatically correct and meaningful sentence.
    \item No information is added or left out, compared to the original question. 
    Therefore, we asked them to use the same phrases, tenses, etc., as much as possible.
\end{enumerate*} 
The same annotation process was repeated for the opposite direction. We showed each \textit{cloze} MCQ to the experts, asking them to convert it to the \textit{normal} formulation. This step added 1,356 new \textit{normal} MCQs and 2,041 new \textit{cloze} formulations, which can be considered equally educationally valuable as their original teacher-generated counterparts.

\subsection{Bloom's Taxonomy Labels}
\label{sec:bloom_taxo}
The Openstax's API offers access to the revised Bloom taxonomy \cite{anderson2001taxonomy} for some questions. The Bloom taxonomy is one of the most recognized cognitive schemes for classifying questions into different levels of complexity, and it is widely used in the development of test items in the educational community 
\cite{downing2006twelve,masapanta2018systematic}. It categorizes questions into six increasing levels of complexity: \textit{Remember}, \textit{Understand}, \textit{Apply}, \textit{Analyze}, \textit{Evaluate}, and \textit{Create}.
Lower levels (\eg \textit{Remember})
are suitable for assessing students' preparation and
comprehension or for reviewing and summarizing content, while higher levels (\eg \textit{Create}) encourage students to think critically and to solve problems \cite{davis2009tools}. 
Among all retained questions after filtering the dataset (see \secref{sec:answer_sentence_selection})
we were able to further enrich 903 questions with the Bloom's taxonomy label. The distribution of questions among the different levels is as follows: 660 in \textit{Remember}, 114 in \textit{Understand}, 110 in \textit{Apply}, and 19 in \textit{Analysis}. No questions for 
the last two categories (\textit{Evaluate} and \textit{Create}) were found, likely due to the fact that we dropped mathematical questions, hypothetical, and opinion-based questions. Although this might seem as a limitation in our study, it is important to note these two categories are less prevalent in current-day factoid-based QG.



\subsection{\dsname{} Statistics}
\tabref{tb:basic_stats} presents some statistics of {\dsname}. The first four rows shows the statistics of entry pairs (normal and cloze), distractors, chapters, and courses in \dsname{}.
The next row (5) counts the questions of the types `none-of-the-above' (NOTA) or `all-of-the-Above' (AOTA).  
We brought this feature up because, in addition to answer's length, it plays an important role in the focusability of MCQs (An MCQ should be answerable without looking at the response options). Although the focusability has received almost no attention from the QG community, it has its own place in educational and teaching research \cite{case1998constructing}. Rows 6 to 8 present the average number of words in answers, questions, and chapters.
Rows 9-13 list the number of questions per Bloom label. 


\begin{table}[h!]
\begin{center}
\footnotesize
\caption{Statistics of \dsname{}.}
\begin{tabular}{lc}
\toprule
\multicolumn{1}{l}{\textbf{Feature}} &
\multicolumn{1}{c}{\textbf{Statistic}} \\
\midrule
1.  \# (normal, cloze) pairs & 3397 \\ 
2.  \# of distractors & 10172 \\
3.  \# of chapters & 241 \\
4.  \# of courses & 12 \\
5.  \# of NOTA and AOTA & 92 \\


6.  Avg. length of answers & 4.1 \\
7.  Avg. length of questions &   12.3 \\
8.  Avg. length of chapters & 12641.5 \\

\midrule
9.  \# of pairs w Bloom's taxonomy & 903 \\
10.  \hspace{1em}   \textit{Remember} & 660 \\
11.  \hspace{1em}  \textit{Understand} & 114 \\
12.  \hspace{1em}  \textit{Apply} & 110 \\
13.  \hspace{1em}  \textit{Analysis} & 19 \\

\bottomrule
\end{tabular}
\end{center}
\label{tb:basic_stats}
\end{table}

\section{Data Analysis}
\label{sec:data_analysis}
We analyze \dsname{} in terms of 
four criteria: the minimum required sentences count, question-context overlap, answer-context overlap, as well as Bloom's labels, and compare it with SQuAD (\tabref{tb:data_analysis}). We chose to use SQuAD not only because it is the most commonly used dataset for QG but also one of the very few datasets where questions and grounding answers are aligned. None of the educational datasets (to our knowledge) provide such a feature.  This alignment limits the variability in generated questions and therefore leads to more accurate automatic evaluation metrics, when comparing with the teachers’ questions.

The \textbf{Minimum required sentences count}, denoted by \mrsc{} in the table, is one of the factors that affect the difficulty of QG and we define the minimum required sentences as the smallest set of reference sentence(s) in the available course material that allow a human to answer the considered question%
\footnote{We do not take into account background knowledge, common reasoning, etc., that the annotators relied on to judge answerability in combination with the selected questions.}.
 We assume it corresponds to the sentences selected by the annotators within the grounding paragraphs (see \secref{sec:answer_sentence_selection}). 
The number of \mrc{} (i.e., \mrsc)
varies from a \textit{single} sentence to \textit{multiple} ones. The questions that rely on multiple sentences appear more thought-provoking to create, compared to the single-sentence ones. 
In our collection, answering 37.6 percent of the questions relies on a \textit{single} sentence, and 62.4 percent on \textit{multiple} one. In SQuAD, all answers are short spans literally mentioned in a \textit{single} sentence, which means its \mrsc{} is 100 percent for \textit{single} cases (see \tabref{tb:data_analysis}). 

The \textbf{Answer-Context overlap} is quantified as the percentage of questions for which the answer literally appears in the \mrc{}. Higher values indicate students may more easily select the correct answer through memorization.
We calculated the lexical match in two ways: \textit{exact} vs. \textit{normalized} (the latter referring to lowercasing, stopword and punctuation removal, stemming, on both answer and \mrc{}). As shown in the table (indicated by Ans-\mrc{}), 38.1 percent of the answers exactly appeared on the \mrc{} in our dataset, and this number increases to 53.9 when we apply the normalization. For SQuAD, this number is 100 percent since by construction, all answers are literally mentioned in the \mrc{}.

The \textbf{Question-Context overlap} is measured as well, assuming that real-world educational questions typically transcend 
a simple syntactic transformation of a declarative sentence \cite{vanderwende2008importance}. The third block in the table (denoted by Que-\mrc{}) presents the results for different levels of \textit{normalized} word overlap (with the same normalization steps as for the normalized Ans-MRS) between the question and \mrc{} (i.e., the number of words they have in common after normalization). For example, 14.5 percent of questions in our dataset have at most one word in common ($0\leq\textit{overlap}\leq1$), compared to is 12.1 percent in SQuAD. Although the gap might seem negligible, we should consider that 62.4 percent of the questions in \dsname{} rely on multiple sentences. The results for high-overlapping cases (\eg $5<\textit{overlap}$) also indicate a slightly higher tendency of crowdworkers to reuse the same words from the context compared to the educators.

\textbf{Bloom's labels} can be a valid proxy to estimate the cognitive level required to answer the question. Therefore this factor can directly contribute to the QG difficulty \cite{pan2019recent}. Not only from the QG perspective but also for learning purposes that is important, since teachers usually combine a mixture of easy and difficult questions to differentiate between weaker or stronger students in the subject. Therefore a suitable educational dataset should be representative of the variant cognitive levels. As presented in the table, most questions (73.1\%, according to the subset of Bloom-labeled questions) in our dataset fall into the \textit{Remember} category, and the rest goes into the 
next three levels. Although that seems out of balance 
at a first glance, it is in line with literature indicating that about 70 percent of asked questions are shallow, and the rest are deep and high-level ones \cite{tofade2013best}. For SQuAD, all question reside in the first level as stated in \cite{chen2018learningq}.

\begin{table}[ht!]
\begin{center}
\footnotesize
\caption{The comparison between our collection and SQuAD with regarding four criteria: Minimum required sentences (\mrc{}), Answer-Context overlap (Ans-\mrc{}), Question-Context overlap (Que-\mrc{}) and Bloom's labels. The \mrc{} stands for the minimum number of sentence(s) needed to answer a question.}
\begin{tabular}{llcc}
\toprule
\multicolumn{1}{l}{\textbf{Property}} & 
\multicolumn{1}{l}{\textbf{Type}} & 
\multicolumn{1}{c}{\textbf{\dsname{}}} &
\multicolumn{1}{c}{\textbf{SQuAD}} \\
\midrule
\mrsc{} & \textit{Single} & 37.6 & 100 \\
& \textit{Multiple} & 62.4 & 0.0 \\
\midrule
Ans-\mrc{} 
& \textit{Exact} & 38.1 & 100 \\
& \textit{Normalized} & 53.9 & 100 \\
\midrule
Que-\mrc{}  
& $0\le\textit{overlap}\le1$  & 14.5 & 12.1 \\
& $2\le\textit{overlap}\le3$  & 45.8 & 41.1 \\
& $4\le\textit{overlap}\le5$  & 27.0 & 31.8 \\
& $5<\textit{overlap}$        & 12.7 & 15.0 \\

\midrule
Bloom's  & \textit{Remember} & 73.1 & 100 \\
label& \textit{Understand} & 12.6 & 0.0 \\
& \textit{Apply} & 12.2 & 0.0 \\
& \textit{Analyze} & 2.1 & 0.0 \\

\bottomrule
\end{tabular}
\end{center}
\label{tb:data_analysis}
\end{table}

\section{MCQ Experiments}
\label{sec:methods_and_experiments}
\label{sec:baseline_models}
We focus our experiments on the creation of MCQ (although the {\dsname} dataset could be used for QA as well). 
We cover the tasks of question generation (\secref{sec:question_generation}), distractor generation (\secref{sec:distractor_generation}), and question format conversion (\secref{sec:format_conversion}).
instructors to switch formats at wish.

For all experiments, we sampled 20\% of chapters along with their questions for testing. We preferred the chapter-level split over randomly sampling questions 
to avoid unwanted and indirect knowledge transfer across questions in the chapter. This division leads to 671 questions for testing and 2,726 for training. Among 671 questions in the test-set, 176 questions have Bloom's taxonomy labels that are distributed as follows: 136 in \textit{Remember}, 18 in \textit{Understand}, 18 in \textit{Apply}, and 4 in \textit{Analyze}. For all generation models, we adopt simple but strong baselines by finetuning the ‘base’ version of T5 \cite{raffel2020exploring} on the target tasks. Further details about the experimental setting and hyperparameters are reported in the appendix (\secref{sec:hyperparameters}).
For the evaluation, we employed standard metrics to evaluate the results: BLEU \cite{papineni2002bleu}, ROUGE$_l$ \cite{lin-2004-rouge}, METEOR \cite{banarjee2005}, token-level F1-score \cite{rajpurkar2016squad}, and Exact Match (EM).

\subsection{Question Generation}
\label{sec:question_generation}
We first investigate the potential of finetuning T5 \cite{raffel2020exploring} for educational QG.
Three fine-tuned versions are evaluated on the {\dsname} test-set: 
\begin{enumerate*}[(i)]
\item T5 finetuned on SQuAD,
\item T5 finetuned on \dsname{},
\item and T5 multi-stage finetuned, first on SQuAD and then on \dsname{} (SQuAD${\rightarrow}$\dsname{}).
\end{enumerate*}
 For all variations, we finetuned T5 in an answer-aware mode where the model not only receives the context but also the target answer as an input to generate the question. Although leaving out the answer may seem more appealing, it could lead to unanswerable questions \cite{sun2018answer}. 
 
 The first block in \tabref{tb:mcq_experiment} presents our QG results. The numbers are in line with earlier studies that reported multi-stage finetuning is beneficial and in-domain finetuning (denoted by \dsname{}) can boost the performance \cite{gururangan2020don, hadifar2021million}. In particular, the multi-stage finetuning (SQuAD${\rightarrow}$\dsname{}) consistently outperforms the two other variants on all evaluation metrics (e.g., by 6.86 and 0.79 BLEU points over SQuAD and \dsname{}, respectivly). Importantly, in-domain finetuning on \dsname{} is  superior to finetuning on the much larger SQuAD dataset. This confirms the complementary value for QG of the new \dsname{} dataset for the educational domain.

 We followed this conventional paradigm to evaluate the quality of generated questions since other techniques, such as asking teachers to review the quality of generated questions or connecting question quality to student performance, are not reproducible.
Alternatively, asking crowdworkers to rate the quality of questions would lead to potential new biases 
\cite{swiecki2022assessment}. Another possible assessment technique is to replace human responses with an artificial crowd of question-answering models. The validity of such an approach would however be extremely reliant on the type of questions and the specific domain.

\begin{figure*}[t!]
    \centering
    \includegraphics[width=\linewidth]{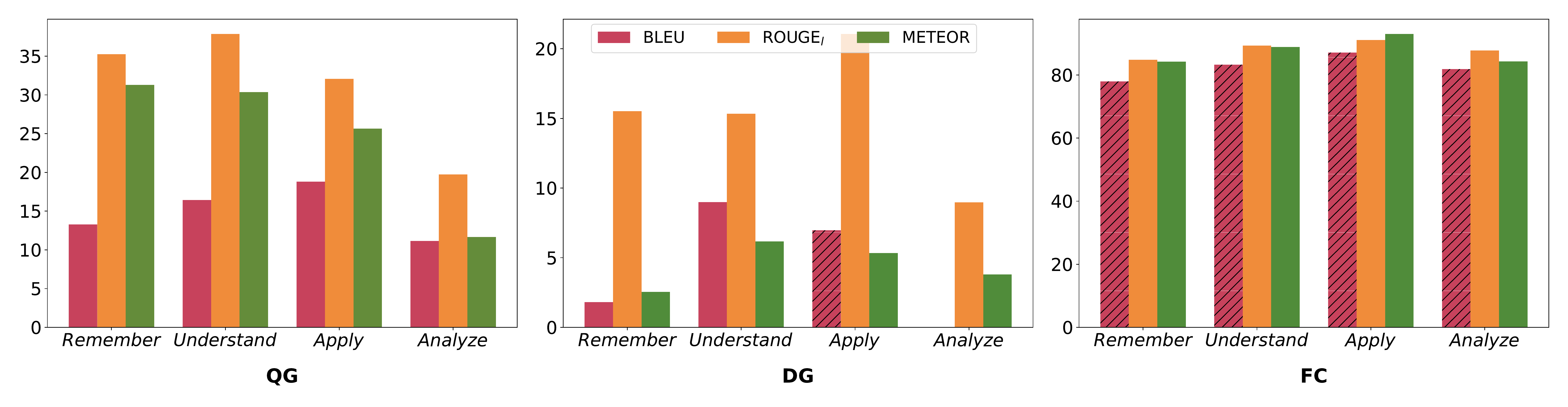}
    \caption{The performance of our models on Question Generation (QG), Distractor Generation (DG), and Format Conversion (FC) for different levels of Bloom's taxonomy.}
    \label{fig:diff_analysis}
\end{figure*}

\begin{table*}[ht!]
\begin{center}
\footnotesize
\caption{MCQ experiments evaluated on the \dsname{} test set. For question generation (QG) and distractor generation (DG), we finetuned T5 in various ways including out-of-domain finetuning (e.g., on SQuAD or RACE), in-domain finetuning (e.g., \dsname{}), and multi-stage finetuning (e.g., SQuAD${\rightarrow}$\dsname{} or RACE${\rightarrow}$\dsname{}). For the format conversion (FC) experiment, two strategies are employed: single-prompt, with T5 finetuned in one direction (e.g, cloze${\rightarrow}$normal), and multi-prompt with T5 finetuned in both directions but tested on the targeted (\underline{underlined}) format (e.g, cloze${\leftrightarrow}$\underline{normal} indicates evaluation on the normal format of the test instances).
}
\begin{tabular}{llccccc}
\toprule
\textbf{Experiment} & 
\textbf{Model} & 
\textbf{BLEU} &
\textbf{ROUGE$_l$} &
\textbf{METEOR} &
\textbf{EM} &
\textbf{F1} \\
\midrule
QG & SQuAD & 8.55 & 24.78 & 20.08 & 0.0 & 25.81 \\
   &\dsname{} & 14.62 &  33.38 & 28.70 & 0.45 & 35.12 \\
   &SQuAD${\rightarrow}$\dsname{} & \textbf{15.41} & \textbf{34.26} & \textbf{29.65} & \textbf{0.75} & \textbf{36.29} \\
\midrule
DG & RACE & 10.88 & 27.50 & 18.10 & 7.28 & 21.45 \\
   & \dsname{} & 17.71 &  32.53 & 20.39 & 9.32 & 24.33 \\
   & RACE${\rightarrow}$\dsname{} & \textbf{17.73} & \textbf{34.13} & \textbf{21.54} & \textbf{10.48} & \textbf{25.74} \\
\midrule
FC & cloze${\rightarrow}$\underline{normal}   & 80.01 & 86.25 & 86.71 & 40.39 & 89.46 \\
  & cloze${\leftrightarrow}$\underline{normal} & 80.26 & 86.38 & 86.95 & 40.69 & 89.71 \\
   & normal${\rightarrow}$\underline{cloze}  & 79.97 & 90.05 & 87.20 & 52.46 & 93.05 \\
  & normal${\leftrightarrow}$\underline{cloze} & 80.24 & 90.36 & 87.36 & 53.80 & 93.17 \\

\bottomrule
\end{tabular}
\end{center}
\label{tb:mcq_experiment}
\end{table*}

\subsection{Distractor Generation}
\label{sec:distractor_generation}
We next look into the automatic generation of distractors. 
We follow the same paradigm as in \secref{sec:question_generation} to build our baselines. Different variants of T5 devised to generate distractors include:
\begin{enumerate*}[(i)]
\item T5 finetuned on RACE \cite{lai2017race},
\item T5 finetuned on \dsname{}, and
\item T5 multi-stage finetuned, first on RACE and then \dsname{} (RACE${\rightarrow}$\dsname{}). \end{enumerate*}
We used parallel decoding, i.e., with all required distractors per question generated in a single decoding step as a \texttt{[SEP]}-separated list. Note that for the reported baselines, we did not evaluate diversity (or lack thereof) of the generated distractors. The results reported in the second block of \tabref{tb:mcq_experiment} are averaged metrics for the different generated distractors, each time evaluated against the entire set of ground truth distractors.

As for the QG task, multi-stage finetuning allows improving upon in-domain finetuning. In-domain finetuning, in turn, 
is superior to the RACE-only model, even though the latter has access to much more training data. 

\subsection{Question Format Conversion}
\label{sec:format_conversion}
The multi-format property of \dsname{} invites new prediction tasks, and in particular
Format Conversion (FC), i.e., the automatic conversion between question formats, in particular from \emph{normal} to \emph{cloze} or vice versa. 
On the one hand, this task is a desirable feature for developers of educational tests. 
On the other hand, it could be 
useful for evaluating masked language models \cite{devlin2018bert} and it could support 
the creation of more challenging datasets \cite{castro2022fiber}. 

Again, we propose different strategies to finetune T5 for this task, with results reported in the third block of \tabref{tb:mcq_experiment}. 
First, we finetuned T5 in a single-prompt fashion, based on one input format, and converting into the other (e.g., `cloze$\rightarrow$normal'). 
Next, in a multiple-prompt experiment, we finetuned conversion in both directions simultaneously.
Test results are reported separately (with the underlined format indicating the evaluation format, as in `cloze${\leftrightarrow}$\underline{normal}' for evaluation on the normal format of the test instances), to allow for comparison with the corresponding single-prompt experiment. %
The multiple-prompt model slightly but consistently outperforms single-prompt finetuning on all evaluation metrics (see generated samples in the appendix \secref{sec:example_predictions}).

\subsection{Question Complexity}
\label{sec:bloom_mcq_analysis}
We further investigated the baseline models in terms of the question complexity. 
We selected the best-performing models in previous experiments for QG, DG, and FC, and evaluate their performance on the Bloom-annotated subset of the test questions. 
As can be seen from \figref{fig:diff_analysis}, increasing levels of difficulty do not necessarily lead to decreased performance for all metrics. For example, the BLEU score for QG as well as DG seems to increase rather than decrease in going from level 1 (\emph{remember}) to level 2 (\emph{understand}). There is a clear decrease in level 4 (\emph{analyze}), though, indicating that
the model struggles to compete with human generated questions and distractors of that level.
We also notice 
that the format conversion results are rather indifferent to the difficulty levels, with high scores for all evaluation metrics. 






\section{Conclusions}
In this paper we introduced \dsname, a new dataset for educational QG based on a collection of OpenStax textbooks, whereby course contents and questions are generated by educational experts. The dataset offers 3,397 high-quality multiple-choice questions, each phrased in a \textit{cloze} as well as \textit{normal} form, and its corresponding answer is linked to the relevant chapter text. Moreover, 903 questions are linked to their cognitive complexity according to Bloom's taxonomy.
We analyzed the data and provided baseline results for question generation, distractor generation, and question format conversion, with clear added value w.r.t.~out-of-domain datasets such as RACE and SQuAD.
We hope \dsname{} will stimulate the development of more advanced teacher-assistant models.

\section*{Acknowledgment}

\newtext{This work was funded by VLAIO (`Flanders Innovation \& Entrepreneurship') in Flanders, Belgium, through the \emph{imec-icon} project AIDA (`AI-Driven e-Assessment').
We would like to thank the partners Televic Education and WeZooz Academy for contributing data and use cases.
}


\ifCLASSOPTIONcaptionsoff
  \newpage
\fi



%

\appendices

\section{List of Study Books}
\label{app:list_of_books}
The following is a list of the books we used as data source for {\dsname}:

\begin{enumerate}
    \item American Government.  \url{https://openstax.org/details/books/american-government-2e}
    \item Anatomy and Physiology. \url{https://openstax.org/details/books/anatomy-and-physiology}
    \item Biology. \url{https://openstax.org/details/books/biology-2e}
    \item Business ethics. \url{https://openstax.org/details/books/business-ethics} 
    \item Business Law i Essentials. \url{https://openstax.org/details/books/business-law-i-essentials}
    \item Intellectual Property. \url{https://openstax.org/details/books/introduction-intellectual-property}
    \item Introduction to Sociology. \url{https://openstax.org/details/books/introduction-sociology-2e}
    \item Microbiology. \url{https://openstax.org/details/books/microbiology}
    \item Financial Accounting. \url{https://openstax.org/details/books/principles-financial-accounting}
    \item Managerial Accounting. \url{https://openstax.org/details/books/principles-managerial-accounting}
    \item Psychology. \url{https://openstax.org/details/books/psychology-2e}
    \item U.S. History. \url{https://openstax.org/details/books/us-history}
    
\end{enumerate}

\section{Annotation Platform}
\label{app:annotation_platform}
\begin{figure}[t!]
    \centering
    
    \includegraphics[width=\columnwidth,angle=270,scale=0.7]{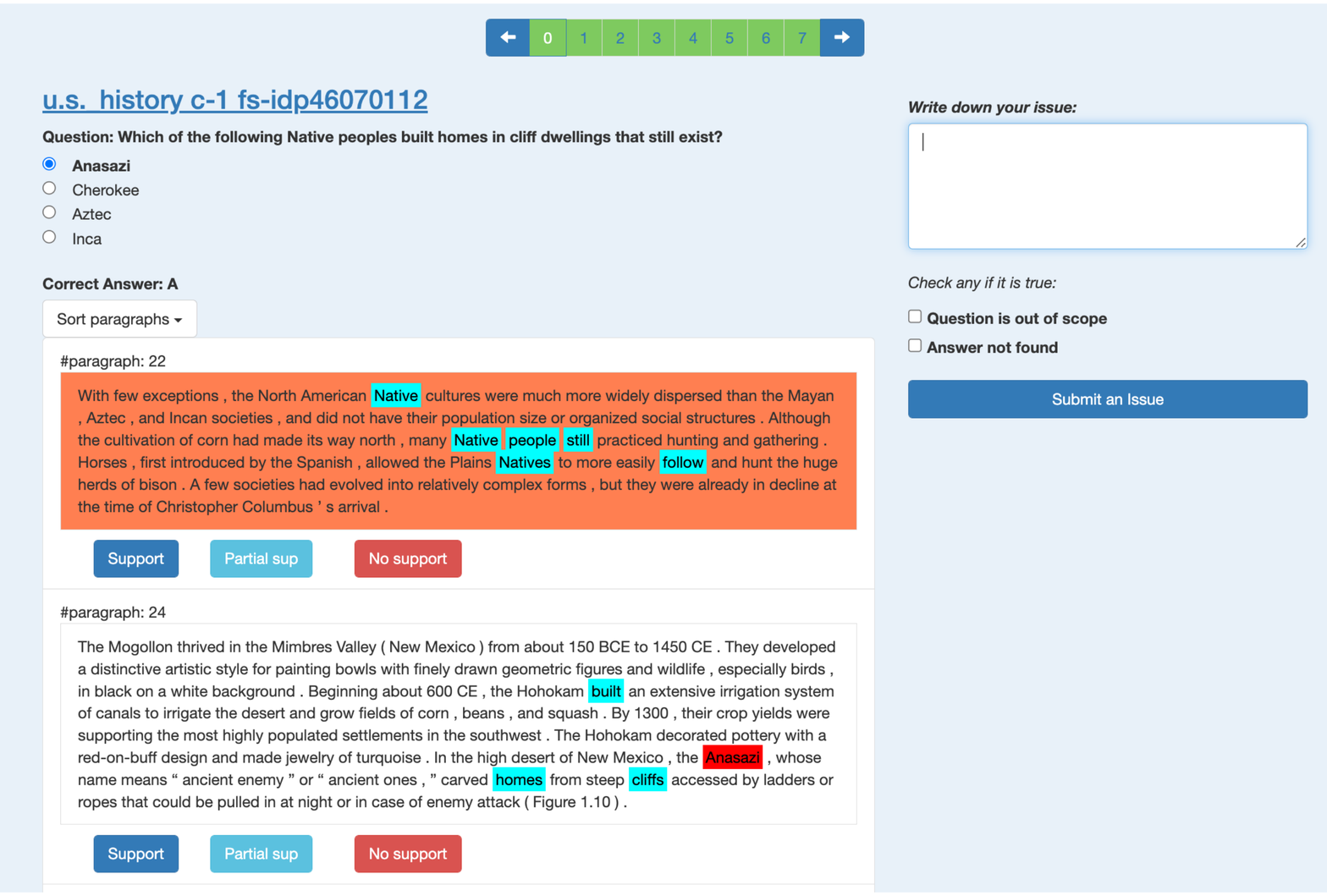}
    \caption{A screenshot of the first stage in our annotation platform. A question, a list of options and aligned paragraphs is shown to the annotator. The annotator should select the label among: support, partial-sup, or no-support.}
    \label{fig:annotation1}
\end{figure}

\begin{figure}[t!]
    \centering
    \includegraphics[width=\columnwidth,angle=270,scale=0.7]{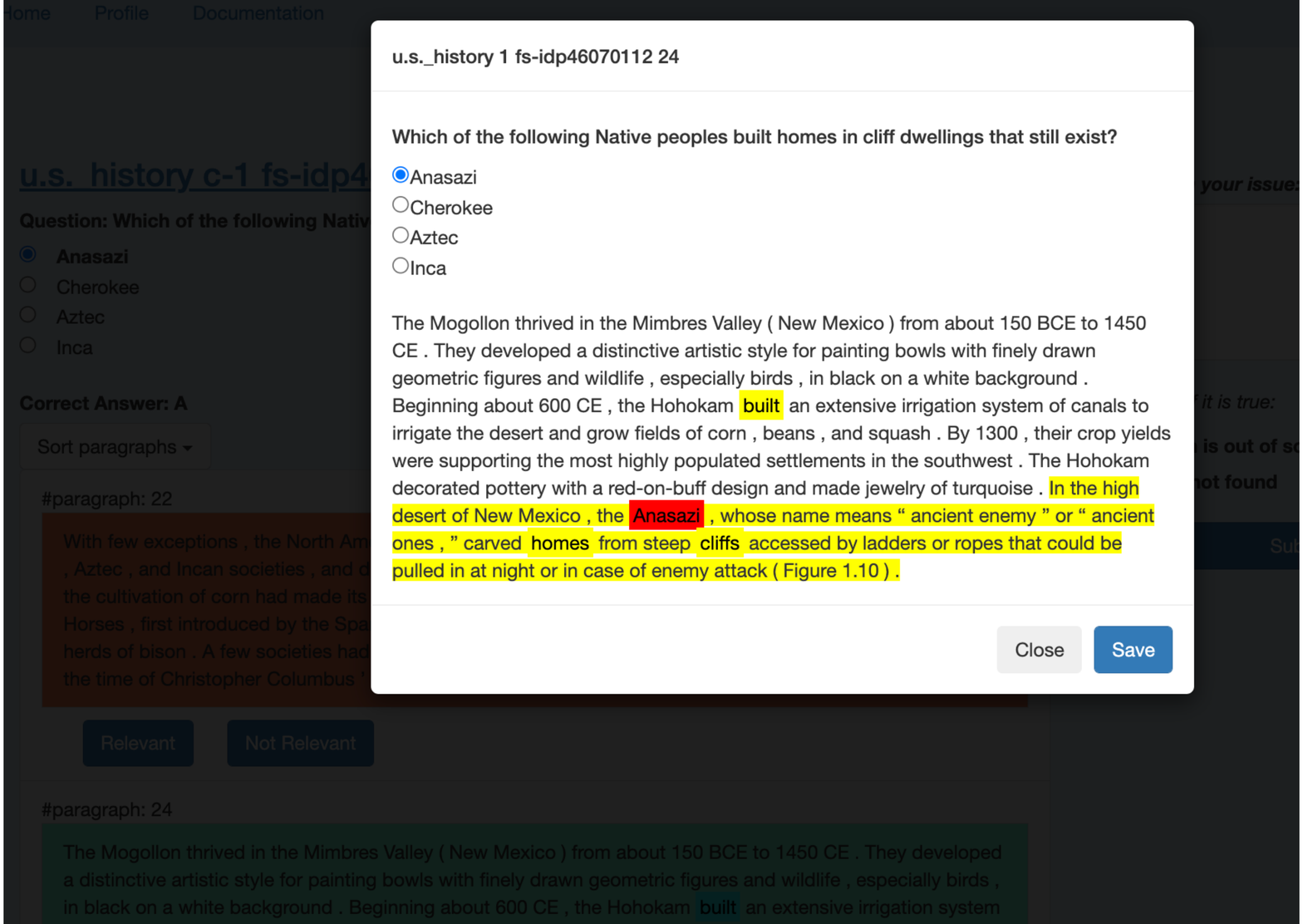}
    \caption{A screenshot of the second stage in our annotation platform. After selecting the paragraph label, the question, the options and the paragraph itself is shown to the annotator for sentence selection. The annotator should select relevant sentence(s) that supports answering the question.}
    \label{fig:annotation2}
\end{figure}

In this section we provide some screenshots of the annotation platform. 
\Figref{fig:annotation1} shows the first stage in the platform. The question, the options (answer in boldface), and a list of paragraphs shown to the annotator. For each paragraph, three button are designed (Support, Partial-support, No-support). In case of any issues, the annotator can submit it in the info-box (in right side of the figure). 
The \figref{fig:annotation2} depicts the second stage of annotation. The annotator enters to the answer selection process after choosing (Support or Partial-support). The selected paragraph along with the question and options is shown to the annotator and they should select the relevant sentence(s) that answer the question.   

\section{Hyperparameters}
\label{sec:hyperparameters}
All experiments were run with one V100 GPU with full precision (FP32) training. T5 was finetuned from pretrained `base' version with the following hyperparameter settings:

\begin{table}[h!]
\begin{tabular}{l}
batch-size =  8 \\
total-epochs =  10 \\
max-source-length = 512 \\
max-target-length = 48 \\
optimizer = AdamW \\
weight-decay = 0.1 \\
adam-epsilon = 1e-08 \\
max-grad-norm = 1.0 \\
lr-scheduler = linear \\
learning-rate = 5e-05 \\
warmup-steps = 500 \\
gradient-accumulation-steps = 4 \\
\end{tabular}
\label{tb:hyperparamters}
\end{table}

\section{Example Predictions}
\label{sec:example_predictions}
In this section, we provide some examples of our best performing models for each of the experiment in \secref{sec:methods_and_experiments}.

\begin{table*}[h!]
    \centering
    \footnotesize
    \caption{Some generated examples from the best performing model in QG, DG, and FC experiments.}
    \begin{tabular}{c p{0.65\linewidth}}
\toprule

\multicolumn{1}{c}{\textbf{Experiment}} & 
\multicolumn{1}{c}{\textbf{Example}} \\

\midrule

      QG
      & 
      \textbf{Input}: competition \texttt{[SEP]} According to sociologist and philosopher Karl Marx, such conflicts are necessary, albeit ugly, steps toward a more egalitarian society. In this pattern of change through conflict, people tend to gain greater personal freedom and economic stability (1848). Philosopher and social scientist Karl Marx was a seminal force in developing the conflict theory perspective. He was a proponent of conflict , in general , because he felt that it was the only means of promoting positive change for the underprivileged. Marx did not agree with Durkheim’s notions of cooperation and interdependence; he instead saw society as a stage for exploitation and strife.      \newline
      \textbf{Ground}: Karl Marx believed social structures evolve out of:     \newline
      \textbf{Prediction}: Marx believed that society should be a stage for what?
      \\
    \midrule
      DG
      & 
      \textbf{Input}: Adams , Murray , and Warren all came from privileged backgrounds . All three were fully literate , while many women in the American republic were not . Their literacy and station allowed them to push for new roles for women in the atmosphere of unique possibility created by the Revolution and its promise of change . Female authors who published their work provide evidence of how women in the era of the American Revolution challenged traditional gender roles . Inspired by the Revolution , Judith Sargent Murray of Massachusetts advocated women \u2019 s economic independence and equal educational opportunities for men and women ( Figure 7.5 ) .  Murray , who came from a well-to-do family in Gloucester , questioned why boys were given access to education as a birthright while girls had very limited educational opportunities . She began to publish her ideas about educational equality beginning in the 1780s , arguing that God had made the minds of women and men equal . Another privileged member of the revolutionary generation , Mercy Otis Warren , also challenged gender assumptions and traditions during the revolutionary era ( Figure 7.5 ) . Born in Massachusetts , Warren actively opposed British reform measures before the outbreak of fighting in 1775 by publishing anti-British works . In 1812 , she published a three-volume history of the Revolution , a project she had started in the late 1770s . By publishing her work , Warren stepped out of the female sphere and into the otherwise male-dominated sphere of public life .  Some women hoped to overturn coverture .  From her home in Braintree , Massachusetts , Abigail Adams ( Figure 7.4 ) wrote to her husband , Whig leader John Adams , in 1776  \texttt{[SEP]} Which of the following figures did not actively challenge the status of women in the early American republic? \texttt{[SEP]} phillis wheatley      \newline
      \textbf{Ground}: ["abigail adams",
            "mercy otis warren",
            "judith sargent murray"
]     \newline
      \textbf{Prediction}:[ "john adams",
            "judith sargent murray",
            "mercy otis warren"]

     \\
     \midrule
      FC
      & 
      
      \textbf{Input}: What the electrons added to NAD+ do, is that \_\_\_.    \newline
      \textbf{Ground}: What do the electrons added to NAD+ do?     \newline
      \textbf{Prediction}: What do the electrons added to NAD+ do?     \newline
      
      \textbf{Input}:  How many NADH molecules are produced on each turn of the citric acid cycle?    \newline
      \textbf{Ground}: The number of NADH molecules that are produced on each turn of the citric acid cycle is \_\_\_.     \newline
      \textbf{Prediction}: On each turn of the citric acid cycle, \_\_\_ NADH molecules are produced. \\
    \bottomrule

    \end{tabular}
    \label{tab:my_label}
\end{table*}

\bibliography{bare_jrnl}
\bibliographystyle{IEEEtran}


\ifCLASSOPTIONcaptionsoff
  \newpage
\fi



%


\begin{IEEEbiography}[{\includegraphics[width=1in,height=1.25in,clip,keepaspectratio]{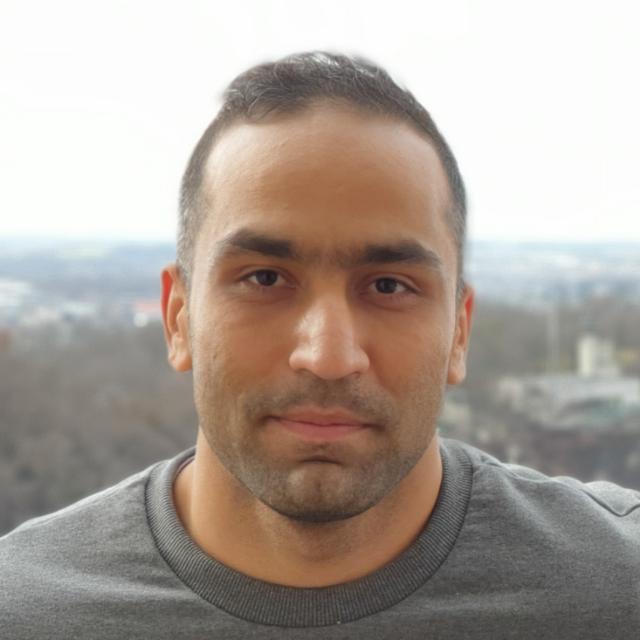}}]
{Amir Hadifar}
is a PhD student at the Internet Technology
and Data Science Lab (IDLab)
at the Ghent University. He is
part of the Text-to-Knowledge
(T2K) Group. His supervisors
are Prof. Chris Develder and Prof.
Thomas Demeester. He received his master’s degree in Computer Engineering from University of Amirkabir Tehran, in 2018. His research interest includes AI in education and conversational systems.
\end{IEEEbiography}

\begin{IEEEbiography}[{\includegraphics[width=1in,height=1.25in,clip,keepaspectratio]{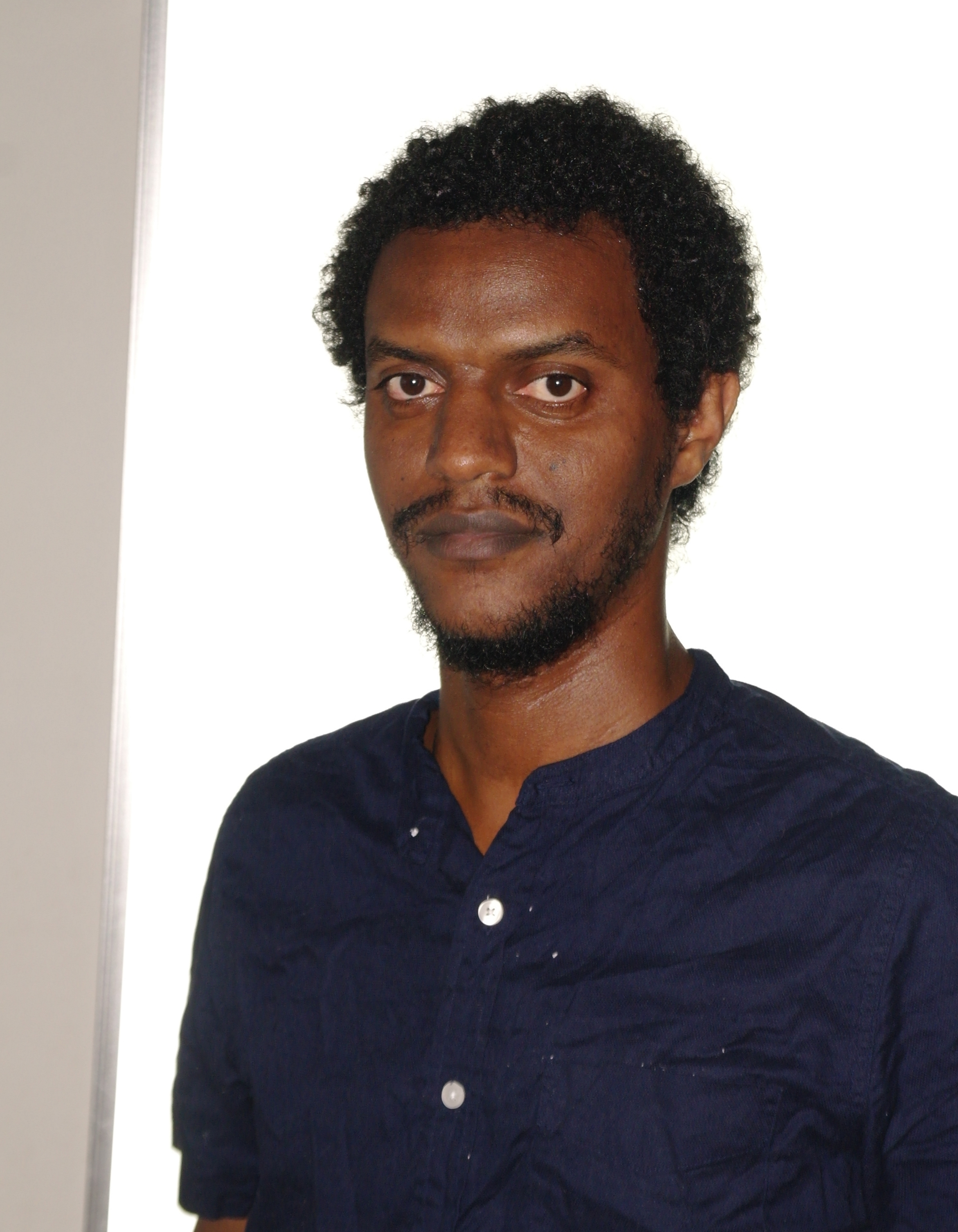}}]%
{Semere Kiros Bitew}
is a PhD student at the Internet Technology
and Data Science Lab (IDLab)
at the Ghent University. He is
part of the Text-to-Knowledge
(T2K) Group. His supervisors
are Prof.\ Chris Develder and Prof.\
Thomas Demeester. He received his master’s degree in Data Science \& Smart services from University of Twente, The Netherlands in 2018. His research interest includes AI in education, controlled text generation and psycho-linguistics. 
\end{IEEEbiography}

\begin{IEEEbiography}[{\includegraphics[width=1in,height=1.25in,clip,keepaspectratio]{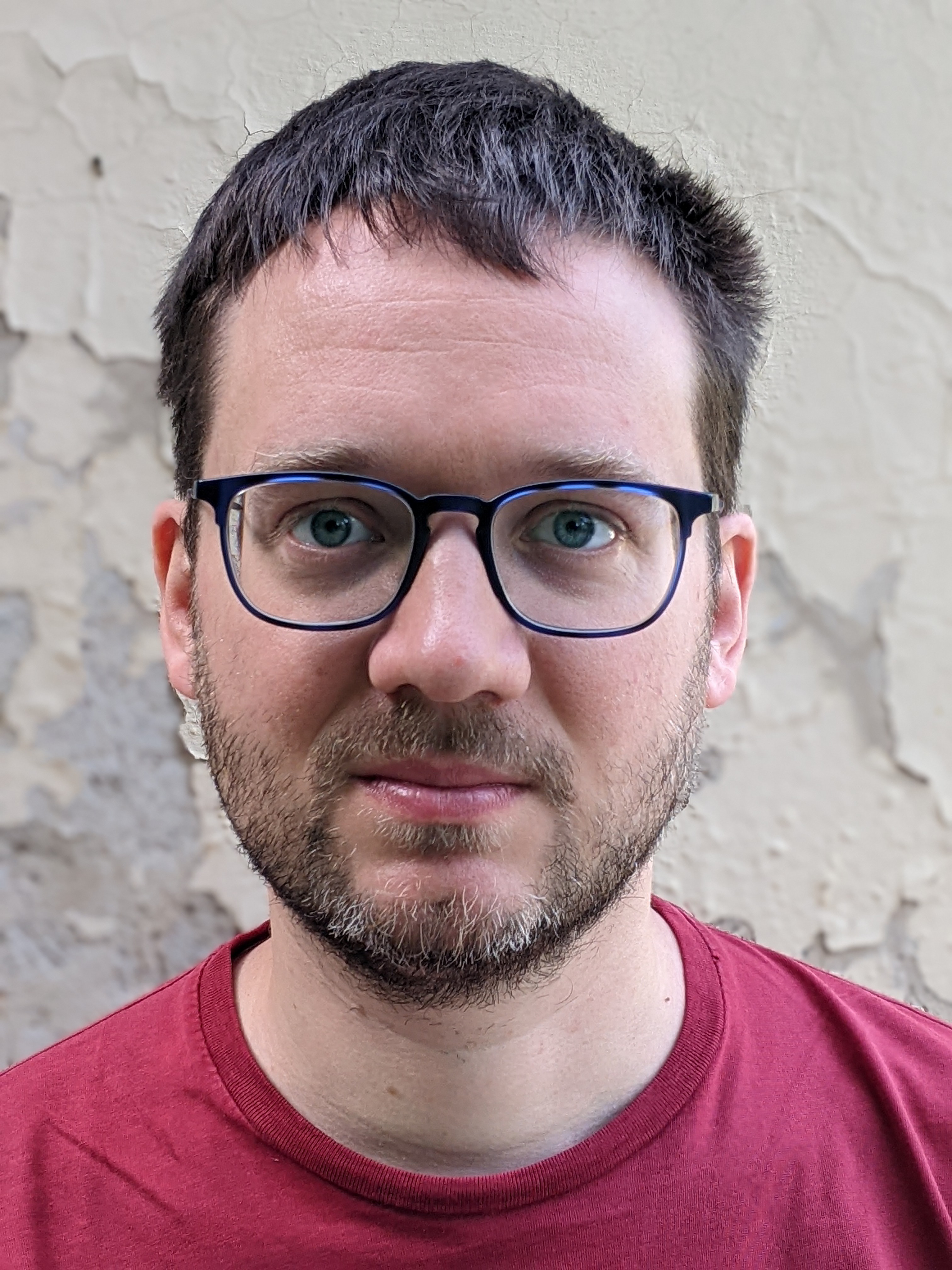}}]%
{Johannes Deleu}
received the
Master of Science degree in
computer science engineering
from Ghent University, Belgium, in 2005. He is currently
a Senior Research Engineer
within the IDLab, Department
of Information Technology,
Ghent University-imec. His
research concentrates on information extraction, machine learning, and in particular deep
learning applied to natural language processing (NLP). He
has participated in multiple
research projects, developing
automatic content enrichment systems for the media sector and more
recently the education sector.
\end{IEEEbiography}

\begin{IEEEbiography}[{\includegraphics[width=1in,height=1.25in,clip,keepaspectratio]{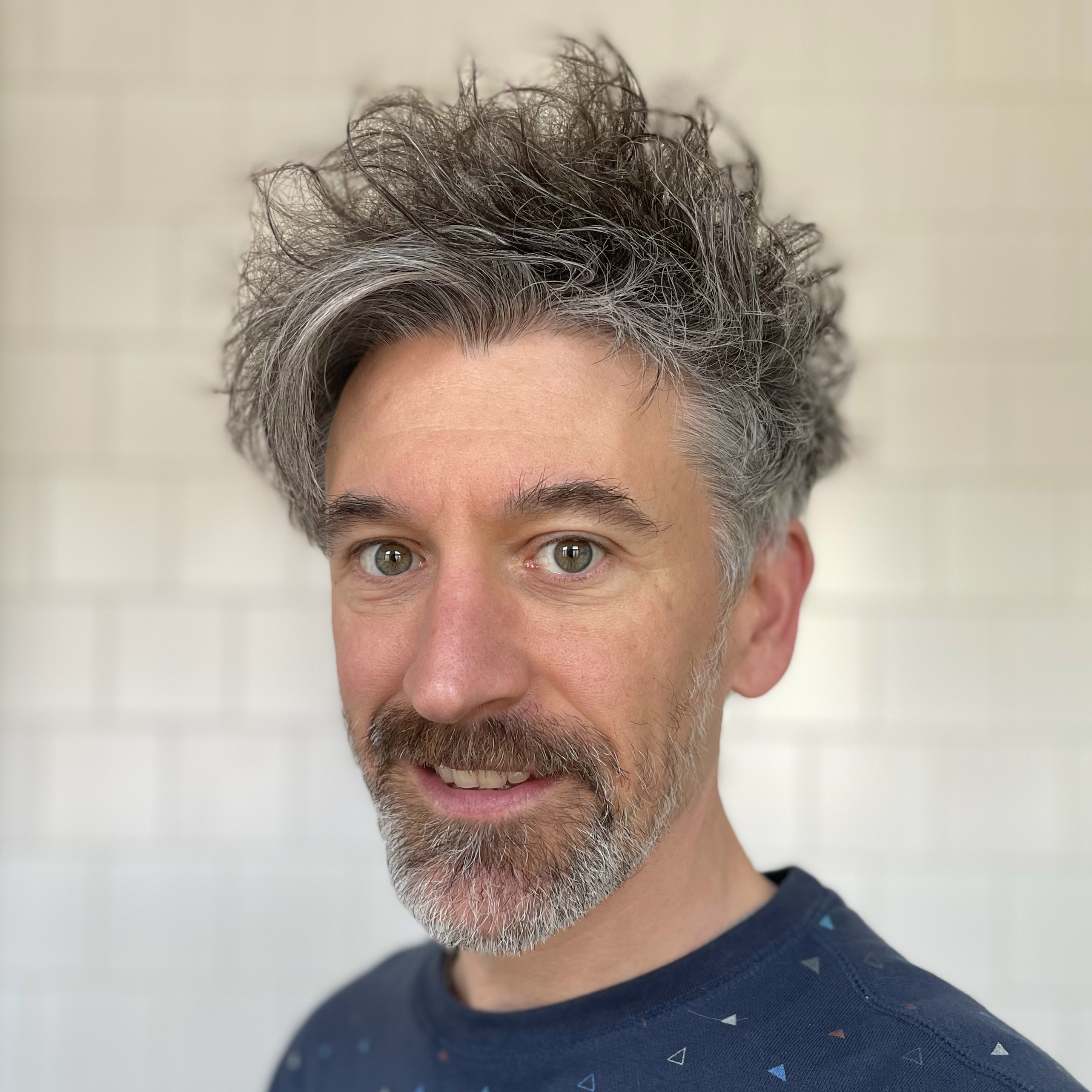}}]%
{Chris Develder}

is associate
professor with the research
group IDLab in the Dept.
of Information Technology
(INTEC) at Ghent Universityimec, Ghent, Belgium. He
received the MSc degree in
computer science engineering and a PhD in electrical
engineering from Ghent University (Ghent, Belgium),
in Jul.\ 1999 and Dec.\ 2003
respectively (as a fellow of the
Research Foundation, FWO).
He has stayed as a research
visitor at UC Davis, CA, USA
(Jul.-Oct.\ 2007) and at Columbia University, NY, USA (Jan.\ 2013 -
Jun.\ 2015). He was and is involved in various national and European
research projects (e.g., FP7 Increase, FP7 C-DAX, H2020 CPN,
H2020 Bright, H2020 BIGG, H2020 RENergetic, H2020 BD4NRG).
Chris currently (co-)leads two research teams within IDLab, (i)~UGent-T2K on converting text to knowledge (i.e., NLP, mostly information extraction using machine learning), and (ii)~UGent-AI4E on
artificial intelligence for energy applications (e.g., smart grid). He has
co-authored over 200 refereed publications in international conferences and journals. He is Senior Member of IEEE, Senior Member of
ACM, and Member of ACL.
\end{IEEEbiography}

\begin{IEEEbiography}[{\includegraphics[width=1in,height=1.25in,clip,keepaspectratio]{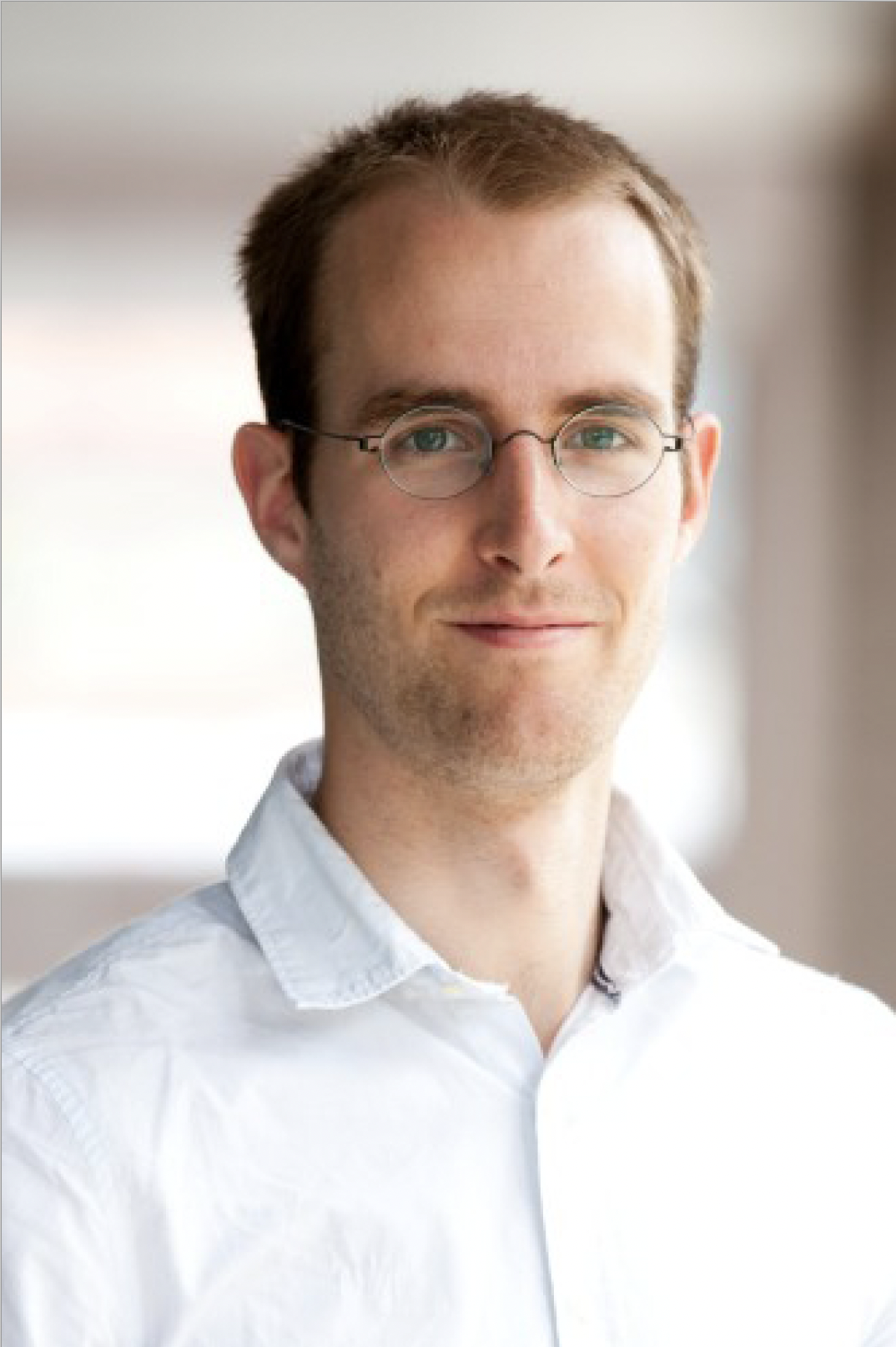}}]%
{Thomas Demeester}
is assistant professor at IDLab, at
the Department of Information Technology, Ghent
University-imec in Belgium.
After his master’s degree in
electrical engineering (2005),
he obtained his Ph.D. in computational electromagnetics,
with a grant from the Research
Foundation, Flanders (FWO-Vlaanderen) in 2009. His
research interests then shifted
to information retrieval (with a
research stay at the University
of Twente in The Netherlands,
2011), natural language processing (NLP) and machine learning (with
a stay at University College London in the UK, 2016), and more
recently to Neuro-Symbolic AI. He has been involved in a series of
national and international projects in the area of NLP, and co-authored
around one hundred peer-reviewed contributions in international
journals and conferences. He is a member of AAAI and ELLIS.
\end{IEEEbiography}
\vfill

\end{document}